# A new ultrasound despeckling method through adaptive threshold

**Hamid Reza Shahdoosti[1,*]**

[1] Department of Electrical Engineering, Hamedan University of Technology, Hamedan, Iran.
email address: h.doosti@hut.ac.ir

**ABSTRACT**

An efficient despeckling method using a quantum-inspired adaptive threshold function is presented for reducing noise of ultrasound images. In the first step, the ultrasound image is decorrelated by an spectrum equalization procedure due to the fact that speckle noise is neither Gaussian nor white. In fact, a linear filter is exploited to "flatten" the power spectral density (PSD) of the ultrasound image. Then, the proposed method shrinks complex wavelet coefficients based on the quantum-inspired adaptive threshold function.

The proposed approach has been used to denoise both real and simulated data sets and compare with other widely adopted techniques. Experimental results demonstrate that the proposed method has a competitive performance to remove speckle noise and can preserve details and textures of medical ultrasound images.

## 1. INTRODUCTION

The random scattering variability in the acquisition procedure of ultrasound images, leads to producing low-contrast and damaged images, typically with a granular surface appearance [1]. This phenomenon is commonly considered as a source of noise, and much effort has been devoted to remove this noise. Speckle noise severely damages the quality of the medical ultrasound image and increases the error of clinical diagnosis and therapy.

Despeckling algorithms of medical ultrasound images have attracted many studies, both in the frequency domain and the spatial domain [2]. With the major limitation of poor edge preservation, the Kuan, Lee and Frost filters are the most classical adaptive filters by which speckle noise is reduced [3, 4]. A very efficient denoising approach is the so called non-local mean (NLM) [5]. In brief, this algorithm assumes that natural images contain a large number of similar regions, commonly referred to as similar patches, positioned at different locations of the image. The GenLik method proposed in Ref. [6] is one of the most successful wavelet-based algorithms [2–4], but the success of despeckling depends on the selection of parameters, and the capability of speckle suppression is limited even when the optimum parameters are selected. Khare et al. [7] introduced a despeckling approach that first detects strong edges using the products of an imaginary part of complex scaling coefficients and then shrinks the coefficients in the wavelet domain at non-edge regions. The non-edge coefficients contain the weak edge information in this approach, owing to the fact that the inter-scale dependency is not considered in the shrinkage function. Offering the properties of approximate shift invariant and directional selectivity, the dual-tree complex wavelet transform is a valuable version of the traditional real discrete wavelet transform [8] by which a better performance can be achieved. Dual-tree complex wavelet transform-based denoising algorithms can reduce the severe artifacts of the critically sampled wavelet transform [9]. Consequently, this transform is employed to despeckle medical ultrasound images in our method.

Some quantum-inspired approaches for ultrasound despeckling have been proposed during last decades. Ref. [10] proposed an adaptive non-Gaussian statistical model for the complex wavelet coefficients. Then, the quantum-inspired probability of noise is determined in the proposed threshold function according to products of the coefficients and coefficients' parents in multi-scale subbands. A novel medical ultrasound despeckling method which combine quantum-inspired bilateral filtering and wavelet thresholding was introduced in Ref. [11]. An adaptive bilateral filter under the concept of quantum signal processing was utilized in this method and exploited as a pre-processing step. Then, the wavelet thresholding function, on the assumption that the signal and noise coefficients in the wavelet domain are subject to generalized Laplace distribution and Gaussian distribution, respectively, is applied to the pre-denoised image.

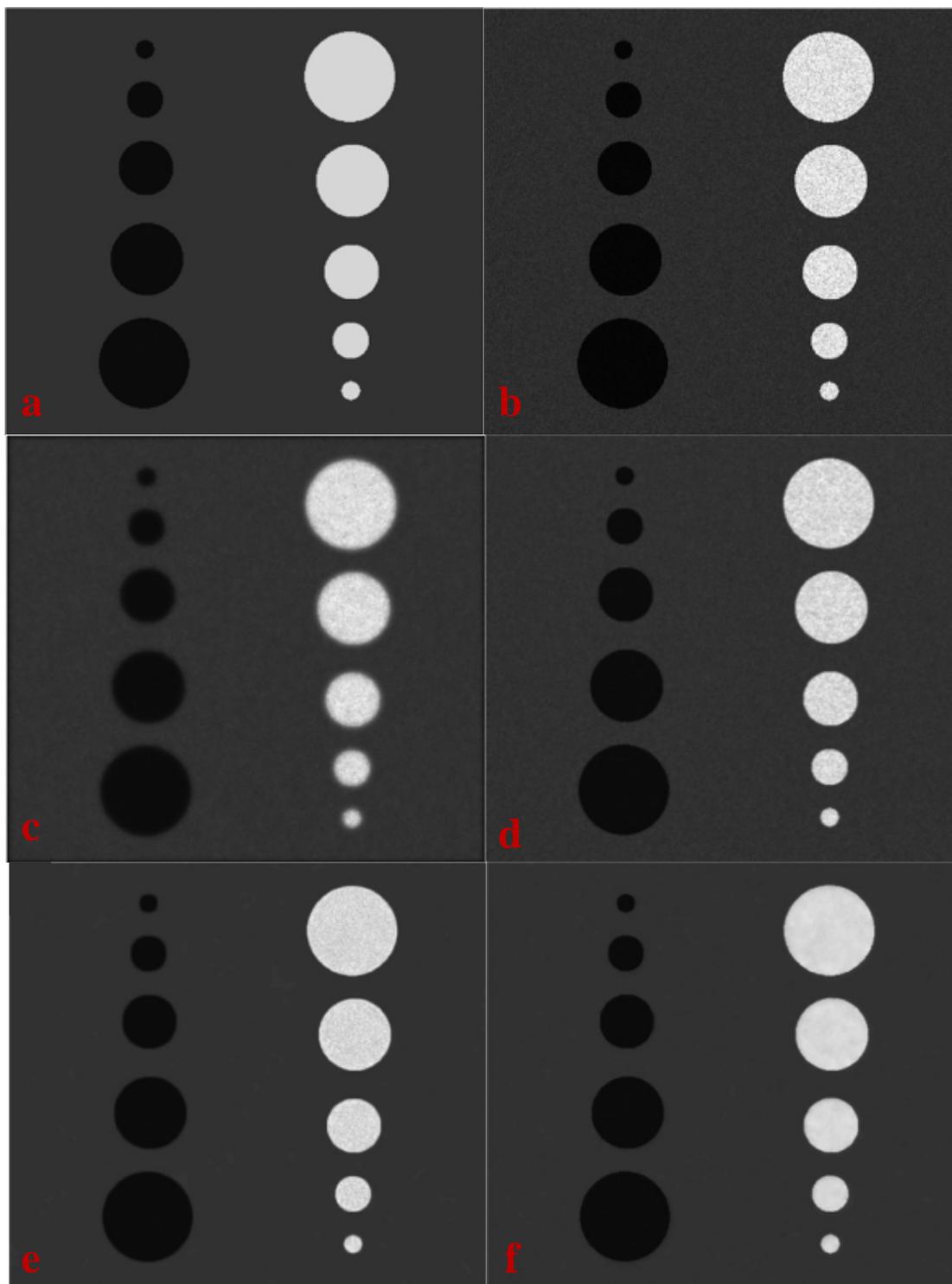

**Fig. 1.** Simulated data. (a) Original data. (b) Noisy data. (c) Frost filter. (d) GenLik method. (e) Log-wavelet based method. (f) The proposed method.

This paper presents an efficient noise filtering method using a quantum-inspired adaptive threshold function for reducing speckle noise of ultrasound images. Firstly, the ultrasound image is decorrelated by an spectrum equalization procedure to change the type of noise into Gaussian and white. In order to broaden the power spectral density (PSD) of the ultrasound image, a linear filter is used. Then, the proposed method denoises complex wavelet coefficients based on the quantum-inspired adaptive threshold function.

The current article is organized as follows. In the next section, the proposed method using a quantum-inspired adaptive threshold function is described. The simulated and real frameworks for performance evaluation, together with comparisons with other widely adopted filters, are presented in section 3. Results achieved and conclusions drawn are covered in the last section.

**2. Proposed algorithm**

*2.1. Preprocessing step*

The speckle noise of ultrasound images can be considered as multiplicative noise [12] (when ignoring the additive component):

$$\tilde{Y} = \tilde{X}\tilde{N} \qquad (1)$$

where $\tilde{Y}$ is the noisy image, $\tilde{X}$ is the original image, and $\tilde{N}$ denotes noise. If a logarithmic transformation is applied to the noisy image, the statistics of speckle noise will alter. The logarithmic transformation affects the high-intensity tail of the noise distribution (here Rayleigh), and thus, the speckle noise becomes closer to Gaussian noise [13]. Then some conventional image restoration methods, such as maximum likelihood filtering, shearlet filtering and anisotropic diffusion, can be applied [14-18]. However, many studies conceptually and experimentally show that this assumption is oversimplified and unnatural [19]. In fact, the speckle noise of ultrasound images is neither Gaussian nor white and thus the preprocessing algorithm proposed in [20] needs to be used to modify the speckle images (without affecting the important information they contain) such that the speckle noise in the log-transformation domain becomes more close to real white Gaussian noise. Then, an effective despeckling algorithm should be employed. To this end, a spectrum equalization procedure is used intending to decorrelate the image samples. A linear filter $l(i, j)$ is designed to broaden the PSD of the image. The transfer function of this filter is as follows:

$$L(w_1, w_2) = (|H(w_1, w_2)| + e)^{-0.5} \qquad (2)$$

where $L(w_1, w_2)$ and $H(w_1, w_2)$ denote the Fourier spectrum of the linear filter and the complex point spread function (PSF) of the envelope detected image, respectively. The free parameter $\varepsilon$ is adjustable and can regulate the decorrelation extent. The magnitude of the spectrum $H(w_1, w_2)$ can be estimated by the method proposed in [20]. Then the shrinkage method described in the following subsection can be applied to the image.

*1.2. Despeckling step*

In the dual-tree complex wavelet transform domain, one can model the log-transformed ultrasound image by $Y = X + N$, where $Y = Y_r + iY_i$, $X = X_r + iX_i$ and $N = N_r + iN_i$ denote the complex wavelet coefficients of $\tilde{Y}$, $\tilde{X}$ and $\tilde{N}$, respectively. Using the MAP estimator, we can write:

$$\hat{X} = \arg\max_X \left(P_{X|Y}(X|Y)\right) = \arg\max_X \left(P_N(N)P_X(X)\right) \qquad (3)$$

Supposing that the distribution of $N$ is close to a Gaussian probability density function (PDF) and using the non-Gaussian statistical model of the complex wavelet coefficients, the following formula is obtained:

$$P_X(X) = \frac{3}{2p\sigma^2}\exp\left(-\frac{\sqrt{3}}{\sigma}\sqrt{X_r^2 + X_i^2}\exp(K)\right) \qquad (4)$$

where $\sigma$ and $K$ are parameters of the model. The MAP estimator of $X$ can be easily derived as:

$$\hat{X} = \frac{\max\left(\sqrt{Y_r^2 + Y_i^2} - \sqrt{3}\exp(K)\sigma_n^2/\sigma, 0\right)}{\sqrt{Y_r^2 + Y_i^2}} \times Y \qquad (5)$$

The magnitude values of high frequency subbands have stronger correlation than those of noise coefficients across diffrent scales. If a parent coefficient of the image is of large magnitude, then its children are more likely to be large. Beacuse the interscale dependency of coefficients exists, the coefficients and their parents are exploited to distinguish between the noise coefficients and signal coefficients in the high frequency scales:

$$C_{xy}^S = |Y^{s+1}(x,y)| \times |Y^s(x,y)| \tag{6}$$

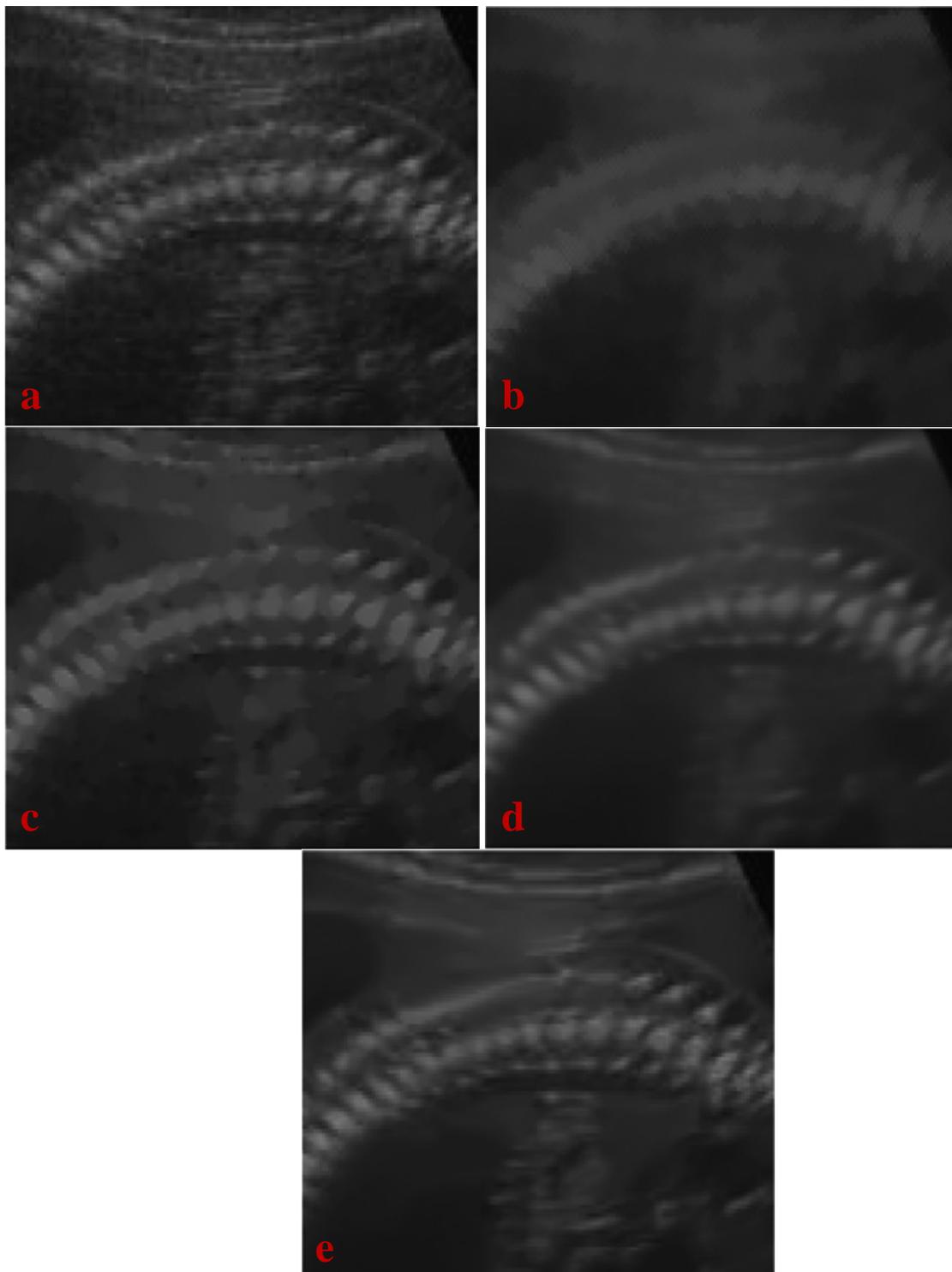

**Fig. 2.** Real ultrasound image. (a) Original data. (b) Frost filter. (c) GenLik method. (d) Log-wavelet based method. (e) The proposed method.

According to the basic concepts of quantum signal processing [21], the state of high frequency scales $C_{xy}^S$ at the $s$th scale can be expressed in a superposition state of noise and image as follows:

$$|C_{xy}^S > = a|0> + b|1> 0 \tag{7}$$

where $a$ and $b$ denote probability of amplitude values of states |0> and |1>, and $|a|^2$ and $|b|^2$ are the measurement probability of noise state |0> and signal state |1>, respectively. In addition, $|a|^2 + |b|^2 = 1$. Let $NC_{xy}^S$ denote the normalized $C_{xy}^S$, reflecting the power of high frequency subbands to some degree. $|C_{xy}^S >$ is defined as:

$$|C_{xy}^S > = \cos(NC_{xy}^S \times p/2)|0> + \sin(NC_{xy}^S \times p/2)|1> \tag{8}$$

The probability of noise $N(m, n)$ in the $s$th level is equal to $\cos^2(NC_{xy}^S \times p/2)$ by which the linear probability of noise is nonlinearly enhanced. So, the parameter $K$ is adaptively determined by $K = \cos^2(NC_{xy}^S \times p/2)$.

The steps of the proposed algorithm are:
1) Apply the spectrum equalization procedure explained in the previous subsection.
2) Calculate the log-transformation of the noisy ultrasound image.
3) Use the complex wavelet to obtain the coefficients of the log-transformed noisy image.
4) Denoise the wavelet coefficients based on the quantum-inspired adaptive threshold function as follows:
   - Estimate the parameter $K$ by $K = \cos^2(NC_{xy}^S \times p/2)$.
   - Update each coefficient using (5).
5) Apply the inverse complex wavelet to the estimated coefficients.
6) Calculate the exponential transformation (inverse of log-transformation) to obtain the denoised image.

### 3. Results

The performance of the proposed quantum-based approach is evaluated on different kind of data sets, both simulated and real ones, and compared with that of other approaches proposed in the literature. Our method is compared with various despeckling techniques such as the Frost filter, GenLik method [6] and the log-wavelet based method [22]. A three level decomposition scheme of complex wavelet transform is used in our approach.

Figs. 1 and 2 show the visual quality of the experiments. Furthermore, table 1 gives the objective evaluation of the experiments, in which the edge preservation measure $\beta$ is used $b \in [0,1]$. The closer the measure $\beta$ approaches to 1, the better the effect of edge preservation is. In addition to the measure $\beta$, the quality of the despeckled images is investigated by the SNR(dB) [23-25]. As can be observed, the proposed method not only gives superior performance in terms of the SNR index, but also has a better edge preservation capacity (see table 1). Moreover, the subjective image quality by the proposed method is much better than the other existing methods in image details preservation and speckle suppression.

**Table 1**. Objective evaluation of different despeckling methods in Figs. 1 and 2.

| Method | Noisy | Frost | GenLik | log-wavelet | Our method |
|---|---|---|---|---|---|
| | | | **Simulated data** | | |
| $\beta$ | 0.463 | 0.427 | 0.571 | 0.726 | 0.872 |
| SNR(dB) | 10.49 | 14.76 | 15.92 | 16.43 | 19.94 |
| | | | **Real data** | | |
| $\beta$ | 0.504 | 0.490 | 0.629 | 0.778 | 0.905 |
| SNR(dB) | 11.76 | 15.33 | 17.02 | 17.59 | 20.31 |

## 4. Conclusions

An effective despeckling method was introduced combining a quantum-inspired adaptive threshold function and an spectrum equalization procedure. We firstly decorrelated the ultrasound image by an spectrum equalization procedure to change the kind of noise into Gaussian white noise. To this end, a linear filter was used to broaden the PSD of the noisy image. Finally, the proposed method denoised complex wavelet coefficients based on the quantum-inspired adaptive threshold function. Experimental results demonstrated that the proposed method has a competitive performance to suppress speckle noise and preserve edges for medical ultrasound images. Furthermore, we showed that this algorithm can provide a feasible despeckling method to improve medical ultrasound image quality by combining basic quantum theory with image processing technology.

## REFERENCES

[1] Baselice, F., 2017. Ultrasound image despeckling based on statistical similarity. Ultrasound in Medicine and Biology, 43(9), pp.2065-2078.
[2] Shahdoosti, H.R. and Rahemi, Z., 2018. A maximum likelihood filter using non-local information for despeckling of ultrasound images. Machine Vision and Applications, 29(4), pp.689-702.
[3] Lopes, A., Touzi, R. and Nezry, E., 1990. Adaptive speckle filters and scene heterogeneity. IEEE transactions on Geoscience and Remote Sensing, 28(6), pp.992-1000.
[4] Dai, M., Peng, C., Chan, A.K. and Loguinov, D., 2004. Bayesian wavelet shrinkage with edge detection for SAR image despeckling. IEEE Transactions on Geoscience and Remote Sensing, 42(8), pp.1642-1648.
[5] Buades A, Coll B, Morel JM. Image denoising methods. a new nonlocal principle. SIAM Rev 2010;52:113–147.
[6] Pizurica, A., Philips, W., Lemahieu, I., and Acheroy, M.: A versatile wavelet domain noise filtration technique for medical imaging', IEEE Trans. Med. Imaging, 2003, 22, (3), pp. 323–331.
[7] Khare, A., Khare, M., Jeong, Y., Kim, H., and Jeon, M.: 'Despeckling of medical ultrasound images using daubechies complex wavelet transform', Signal Process., 2010, 90, (2), pp. 428–439.
[8] Selesnick, I.W., Baraniuk, R.G. and Kingsbury, N.C., 2005. The dual-tree complex wavelet transform. IEEE signal processing magazine, 22(6), pp.123-151.
[9] Shahdoosti, H.R. and Hazavei, S.M., 2017. Image denoising in dual contourlet domain using hidden Markov tree models. Digital Signal Processing, 67, pp.17-29.
[10] Fu, X.W., Ding, M.Y. and Cai, C., 2010. Despeckling of medical ultrasound images based on quantum-inspired adaptive threshold. Electronics letters, 46(13), pp.889-891.
[11] Fu, X., Wang, Y., Chen, L. and Dai, Y., 2015. Quantum-inspired hybrid medical ultrasound images despeckling method. Electronics Letters, 51(4), pp.321-323.
[12] Aubert, G. and Aujol, J.F., 2008. A variational approach to removing multiplicative noise. SIAM journal on applied mathematics, 68(4), pp.925-946.
[13] V. Dutt, Statistical analysis of ultrasound echo envelope, Ph.D. thesis, Mayo graduate school, Rochester, MN, 1995.
[14] Shahdoosti, H.R. and Khayat, O., 2016. Image denoising using sparse representation classification and non-subsampled shearlet transform. Signal, Image and Video Processing, 10(6), pp.1081-1087.
[15] Shahdoosti, H.R. and Hazavei, S.M., 2018. Combined ripplet and total variation image denoising methods using twin support vector machines. Multimedia Tools and Applications, 77(6), pp.7013-7031.
[16] Shahdoosti, H.R., 2017. Two-stage image denoising considering interscale and intrascale dependencies. Journal of Electronic Imaging, 26(6), p.063029.
[17] Khayat, O., Razjouyan, J., Aghvami, M., Shahdoosti, H.R. and Loni, B., 2009, March. An automated GA-based fuzzy image enhancement method. In Computational Intelligence for Image Processing, 2009. CIIP'09. IEEE Symposium on (pp. 14-19). IEEE.
[18] Khayat, O., Shahdoosti, H.R. and Khosravi, M.H., 2008, February. Image classification using principal feature analysis. In Proceedings of the 7th WSEAS International Conference on Artificial intelligence, knowledge engineering and data bases (pp. 198-203). World Scientific and Engineering Academy and Society (WSEAS).
[19] Lai, D., Rao, N., Kuo, C.H., Bhatt, S. and Dogra, V., 2009, June. An ultrasound image despeckling method using independent component analysis. In Biomedical Imaging: From Nano to Macro, 2009. ISBI'09. IEEE International Symposium on (pp. 658-661).
[20] O. V. Michailovich and A. Tannenbaum, "Despeckling of medical ultrasound images," IEEE Transactions on Ultrasonics, Ferroelectrics and Frequency Control, vol. 53(1), pp. 64–78, Jan. 2006.
[21] Eldar, Y.C., and Oppenheim, A.V.: 'Quantum signal processing', IEEE Signal Process. Mag., 2002, 19, (6), pp. 12–32.

[22] Sendur, L., and Selesnick, I.W.: 'Bivariate shrinkage functions for wavelet based denoising exploiting interscale dependency', IEEE Trans. Signal Process., 2002, 50, (11), pp. 2744–2756.
[23] Shahdoosti, H.R., 2017. Robust non-local means filter for ultrasound image denoising. arXiv preprint arXiv:1710.01245.
[24] Shahdoosti, H.R., 2018. Image denoising through bivariate shrinkage function in framelet domain. arXiv preprint arXiv:1801.00635.
[25] Shahdoosti, H.R. and Salehi, M., 2017. Transform-based watermarking algorithm maintaining perceptual transparency. IET Image Processing, 12(5), pp.751-759.